\documentclass{bmvc2k}
\usepackage{graphicx}
\usepackage{amsmath}
\usepackage{enumerate}
\usepackage{txfonts}
\usepackage[linesnumbered,lined,ruled,commentsnumbered]{algorithm2e}
\usepackage{hyperref}
\usepackage{multirow}
\usepackage{amsfonts}

\title{Each Attribute Matters: Contrastive Attention for Sentence-based Image Editing}

\addauthor{Liuqing Zhao *}{liuqingzhao@post.usts.edu.cn}{1}
\addauthor{Fan Lyu *}{fanlyu@tju.edu.cn}{2}
\addauthor{Fuyuan Hu }{ fuyuanhu@mail.usts.edu.cn}{1}
\addauthor{Kaizhu Huang }{kaizhu.huang@xjtlu.edu.cn}{3}
\addauthor{Fenglei Xu }{xufl@mail.usts.edu.cn}{1}
\addauthor{Linyan Li $\dagger$}{lilinyan@szjm.edu.cn}{4}

\addinstitution{
 Suzhou University of\\
 Science and Technology,\\
 Suzhou, China
}
\addinstitution{
 College of Intelligence and Computing,\\
 Tianjin University\\
 Tianjin, China
}
\addinstitution{
 Xi'an Jiaotong-Liverpool University,\\
 Suzhou, China
}
\addinstitution{
 Suzhou Institute of Trade and Commerce,\\
 Suzhou, China
 \\ 
 \rule{\linewidth}{0.4pt}
 \\
 \footnotesize*L.Zhao and F.Lyu share equal contribution.\\
 $\dagger$ L.Li is the corresponding author.\\
}

\runninghead{ZHAO, ET AL.}{Each Attribute Matters}

\def\eg{\emph{e.g}\bmvaOneDot}

\def\etal{\emph{et al}\bmvaOneDot}

\def\mb{\mathbf}
\def\mbb{\mathbb}
\def\mc{\mathcal}

\def\etal{{\em et al.\/}\, }

\def\ie{\textit{i.e.}}

\begin{document}

\maketitle

\begin{abstract}
Sentence-based Image Editing (SIE) aims to deploy natural language to edit an image. Offering potentials to reduce expensive manual editing, SIE has attracted much interest recently.
However, existing methods can hardly produce accurate editing and even lead to failures in attribute editing when the query sentence is with multiple editable attributes.
To cope with this problem, by focusing on enhancing the difference between attributes, this paper proposes a novel model called Contrastive Attention Generative Adversarial Network (CA-GAN),  which is inspired from contrastive training.
Specifically, we first design a novel contrastive attention module to enlarge the editing difference between random combinations of attributes which are formed during training. We then construct an attribute discriminator to ensure effective editing on each attribute.  
A series of experiments show that our method can generate very encouraging results in sentence-based image editing with multiple attributes on CUB and COCO dataset.
Our code is available at \url{https://github.com/Zlq2021/CA-GAN}

\end{abstract}


\section{Introduction}

As billions of images are uploaded and shared every day~\cite{126amazing, rupprecht2018guide}, image editing has become one of the most demanding tasks in social media.
However, to edit an image as desired, one may have to master professional software such as Adobe PhotoShop.
In contrast with manual editing,  automatic image editing,  has recently attracted much interest in computer vision.
This paper studies the problem of Sentence-based Image Editing (SIE)~\cite{dong2017semantic, nam2018text, li2020manigan} that intends to deploy natural language to assist image editing automatically. 
One main challenge for SIE is to build the cross-modal mapping from the query sentence to the pixels in image.
In the last decade, Deep Neural Networks~\cite{lyu2020multi,lyu2019attend} enabling generative models to produce pixel-level manipulation from another image have become the main solution to SIE.


\begin{figure}[t]
	\centering
	\includegraphics[width=\linewidth]{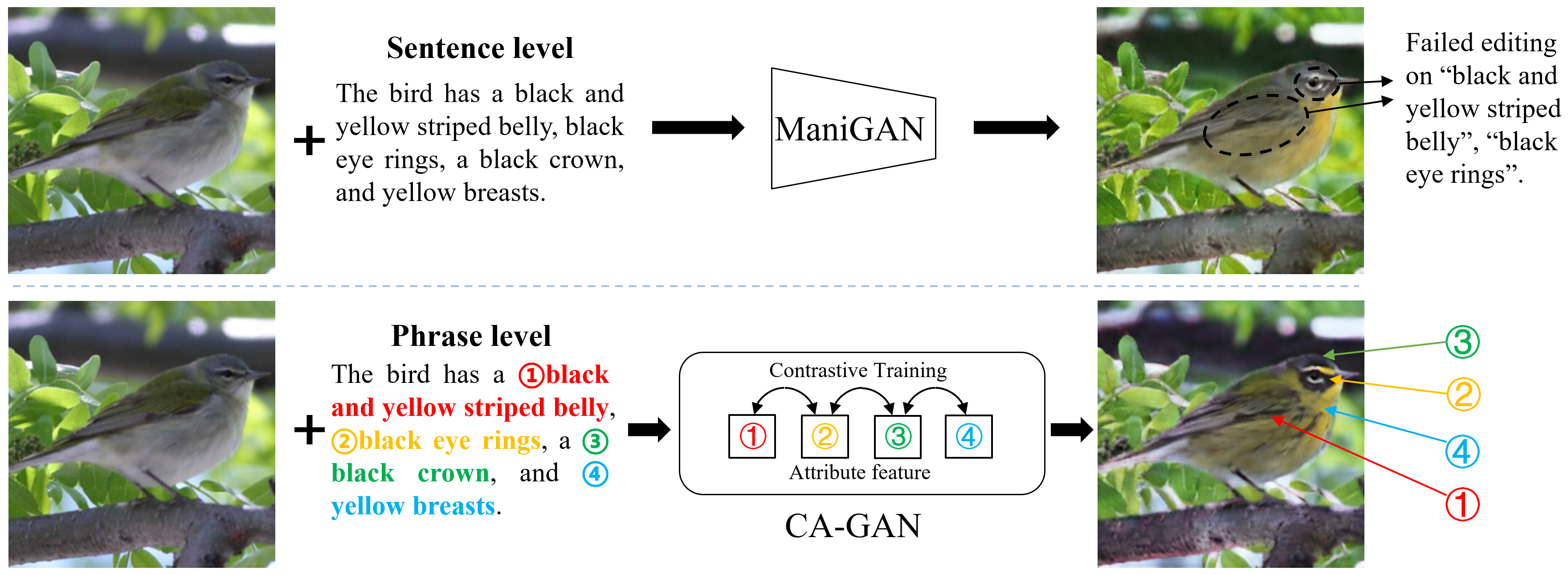}
	\caption{
		Comparisons between the existing sentence-level editing and the proposed Contrastive Attention GAN.
		The state-of-the-art ManiGAN~\cite{li2020manigan} cannot effectively parse different attributes from the given sentence, which yields the failed attribute editing on ``black eye rings''.
		The proposed CA-GAN parses the sentence, learns to distinguish attributes from each other and edits successfully on all attributes.
	} 
	\label{fig1}
	\vspace{-10px}
\end{figure}

Based on Generative Adversarial Networks (GANs)~\cite{zhu2019dm, zhang2018photographic, yin2019semantics, tan2019semantics, qiao2019learn, lao2019dual}, recent works on SIE focus on combining sentence and image information. 
For example, AttnGAN~\cite{xu2018attngan} maps the query sentence and the image to be edited into a shared hidden space and minimizes the multi-modal similarity to improve the quality of text-to-image generation. 
TAGAN \cite{nam2018text} provides word-level feedback to the generator through a fine-grained text discriminator. 
ManiGAN \cite{li2020manigan} combines language and image with a three-stage network structure, which progressively generates images from three different scales. 
These methods show impressive results with short query sentence or phrase.

Nevertheless, when the query sentence is long and contains multiple attributes to be edited, the existing methods can hardly produce effective editing for all attributes.
As a typical example shown in Fig.~\ref{fig1}, a query sentence ``\textit{The bird has a black and yellow striped belly, black eye rings, a black crown, and yellow breasts}'' is used to guide the editing on a given bird image.
Intuitively, the sentence has a few different editable attributes: ``\textit{black and yellow striped belly}'', ``\textit{black eye rings}'', ``\textit{black crown}'' and ``\textit{yellow breasts}''. 
The current state-of-the-art ManiGAN~\cite{li2020manigan} fails to edit the attribute ``\textit{black eye rings}''.
The main reason for the failure is that the existing methods only focus on sentence-level editing rather than each attribute.
Going further, we argue that there are  three main obstacles for these GAN-based sentence-level editing methods:
1) they cannot parse sentences  effectively and the attribute differences are indistinguishable;
2) they cannot build the attribute-pixel correspondence  properly;
3) the sentence-level discriminator  utilised in these methods is limited in detecting the failed attribute editing.

To tackle these drawbacks, we aim to  strengthen each editable attribute so as to attain an accurate SIE model.
Concretely, inspired by the contrastive training, we propose a novel Contrastive Attention Generative Adversarial Network (CA-GAN) for SIE. Our proposed CA-GAN contains three main components:
1) \textbf{Sentence Parsing and Attribute Combination}.
To facilitate the training for the attribute-level editing, we first parse the query sentence based on POS Tagging to ensure the attribute-object correspondence.
Then we augment the query space by random attribute combinations, which prove to  significantly highlight attribute-level information.
2) \textbf{Contrastive Training using attention}.
Intuitively, based on the augmentation from random attribute combinations, different combinations yield different editing.
Thus, we construct the Contrastive Attention for different combinations in the GAN architecture.
Our model can then enlarge the editing difference between any two attribute combinations, whilst keeping the background invariant.
3) \textbf{Attribute-level Discriminator}. 
In the discriminator, we build an attribute-level discriminator for providing effective editing feedback on each attribute to the generator.
With the proposed CA-GAN, the editable attributes in the sentence can be well distinguished via training. Thus an effective SIE model can be generated which emphasizes each attribute appropriately.
To the best of our knowledge, this is the first work that proposes to separate attributes from long sentences to strengthen the attribute editing. 
We evaluate the proposed CA-GAN on two benchmark image editing datasets, \ie, CUB and MS-COCO.
The evaluation results show that our method can edit the attributes at the pixel level effectively  and accurately.

\section{Related Work}


\noindent
\textbf{Sentence-based image editing}.
In recent years, based on Generative Adversarial Networks (GANs)~\cite{zhu2019dm,zhang2018photographic,yin2019semantics,tan2019semantics,qiao2019learn,lao2019dual}, researchers pay much attention to the image generation or transformation from text or image, such as Text-to-Image Generation~\cite{nguyen2017plug,xu2018attngan,qiao2019mirrorgan,zhang2017stackgan,zhang2018stackgan++} and Image-to-Image Translation~\cite{ye2019unsupervised,huang2018multimodal,press2020emerging,yu2018singlegan}.
To make the transformation controllable, Text-based Image Editing will only edit the target area of the image through text description. 
Generally, the query text can be a word, a short phrase or a long sentence.
Dong \etal proposed an encoder-decoder structure to edit images matched with a given text~\cite{dong2017semantic}. 
In order to keep the content irrelevant to the text in the original image,~\cite{vo2018paired} proposed to construct foreground and background distribution with different recognizers. 
Nam \etal eliminated different visual attributes by introducing a text adaptive discriminator, which can provide more detailed training feedback to the generator~\cite{nam2018text}. 
Li \etal adopted the structure of the multi-level network, and could generate high-quality image content through the combination module of ACM and DCM~\cite{li2020manigan}. 
However, the generators of these methods ignore the difference between long sentences with words and phrases that  may contain multiple editable attributes as well.
In this paper, we focus on the Sentence-based Image Editing, and propose to construct Contrastive Attention to enhance the attribute editing.

\noindent
\textbf{Contrastive training}.
For a given anchor point in the data, the purpose of contrast learning~\cite{yu2019multi,he2020momentum,chen2020simple} is to bring the anchor point closer to the positive point and push the anchor point further away to the negative point in the representation space, thus enhancing the consistency of the feature representation. 
In previous vision tasks~\cite{wu2018unsupervised,lee2021infomax,deng2020disentangled,kang2020contragan}, the idea of contrast learning is also applied by exploring the relationship between positive and negative samples. 
It was also demonstrated in~\cite{park2020contrastive} that contrast learning methods are effective in the task of image to image conversion.
Some works also studied the contrastive training in natural language processing~\cite{zhang2020unsupervised, he2020momentum}.
CDL-GAN~\cite{zhou2021cdl} add Consistent Contrastive Distance (CoCD) and Characteristic Contrastive Distance (ChCD) into a principled framework to improve GAN performance.
CERT~\cite{fang2020cert} uses back-translation for data augmentation. 
BERT-CT~\cite{carlsson2020semantic} uses two individual encoders for contrastive learning.
In this work, we argue that different attributes should be edited discriminatively and this leads to the idea of contrast attention on SIE. 
\section{Method}

\begin{figure}
	\includegraphics[width=\textwidth]{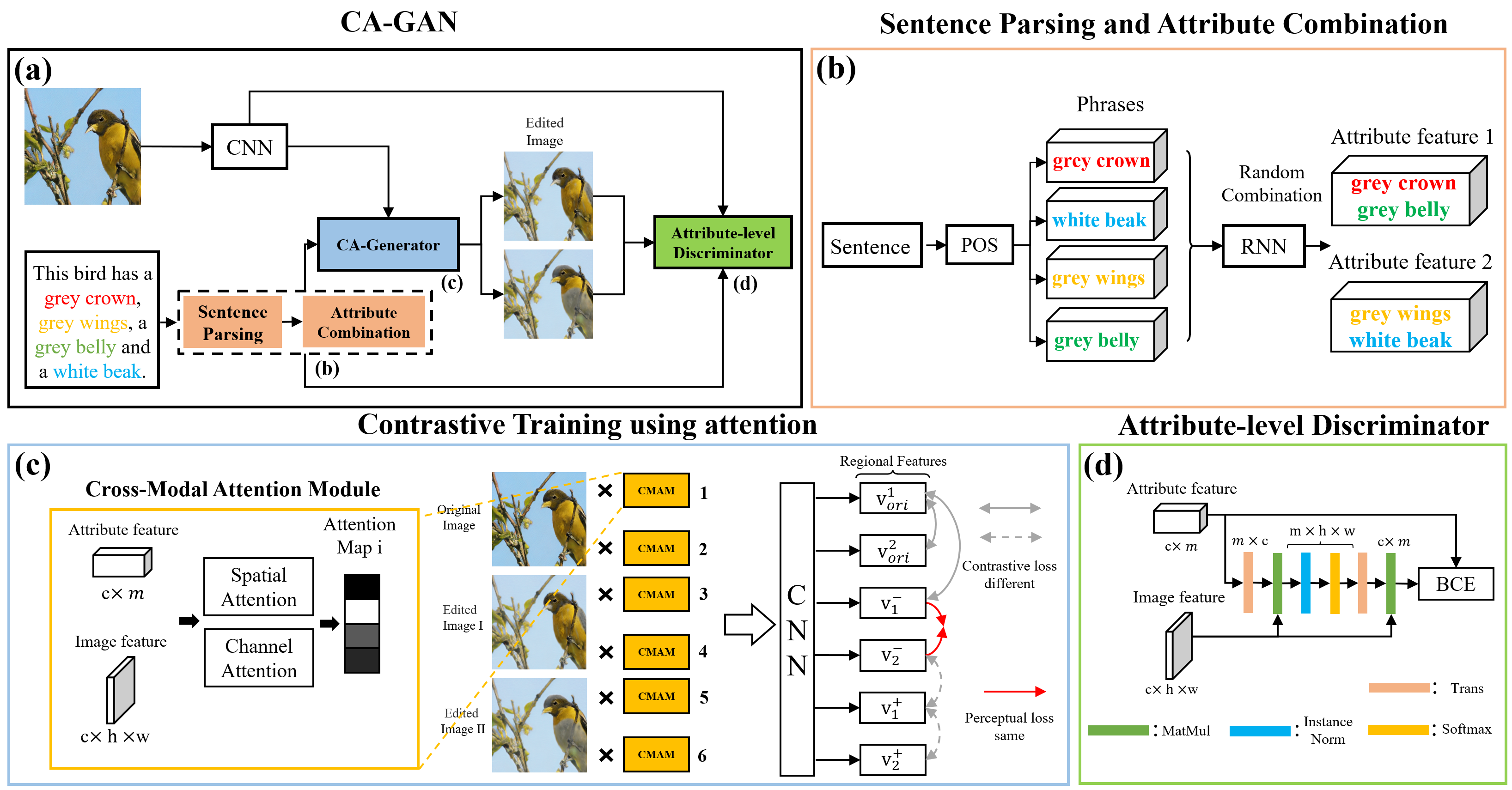}
	\caption{(a) Overview of the proposed CA-GAN. 
		Sentence information is passed through (b), segmentation of editable attributes by sentence analysis, random combination for data augmentation.
		(c) Contrastive Training using attention module. The contrastive training between different attributes is constructed by editing images with different combinations of attributes.
		CMAM module outputs attention maps for different attributes.
		(d) Attribute-level Discriminator, which provides attribute-level feedback to the generator.} \label{fig2}
	\vspace{-15px}
\end{figure}

\subsection{Overview}
Given an image $\mb{I}\in\mbb{R}^{c\times h\times w}$ and a query sentence $\mc{S}$, SIE aims to transform $\mb{I}$ guided by $\mc{S}$ to an edited image $\hat{\mb{I}}$. 
Our Contrastive Attention Generative Adversarial Network (CA-GAN) is based on the popular three-stage editing architecture~\cite{xu2018attngan}.
To be more specific, there are usually three stages in the main module, and each stage contains a generator and a discriminator.
Three stages are trained at the same time, and progressively generate images of three different scales, i.e., $64^{2}$→ $128^{2}$→ $256^{2}$.
As shown in Fig.~\ref{fig2}, CA-GAN contains three main components:
(1) \textbf{Sentence Parsing and Attribute Combination} (Fig.~\ref{fig2}(b)). The query sentence is parsed to multiple editable attributes based on a Lexical rules based on POS tagging~\cite{bird2009natural}. Then, the attributes are randomly combined into two groups for augmentation.
(2) \textbf{Contrastive Training using attention} (Fig.~\ref{fig2}(c)). This module uses attention distinguishes different combinations, and each attribute can be learned.
(3) \textbf{Attribute-level Discriminator} (Fig.~\ref{fig2}(d)). This module is designed to provide the feedback if an attribute is edited well.
We will elaborate these steps in the following.


\subsection{Sentence Parsing and Attribute Combination}\label{3.2}
Some off-the-shelf methods can parse a sentence to multiple phrases, such as Topicrank~\cite{bougouin2013topicrank} and Sentence Transformers~\cite{nikzad2021phraseformer} and BERT~\cite{devlin2018bert}.
However, these methods suffer from two main problem:
1) task-specific design and not for image editing;
2) large-scare network with massive parameters.
In contrast, we propose a parsing rule to effectively parse attributes from a sentence and perform in lightweight scale.
Specifically, given a sentence $\mc{S}$, as shown in Fig.~\ref{fig2}(c), we first use POS tagging~\cite{bird2009natural} (NLTK~\cite{loper2002nltk}) to label each word in a sentence with a lexical property (\eg noun, adjective).
With the lexical property, we need to further separate the attributes from the sentences in the form of ``adjective-noun''.
But it is difficult to determine the adjectives belongingness of two neighboring attributes in the sentence. 
This can be illustrated in the examples  ``\textit{bird with a black wing}''$\rightarrow$``[\textit{bird black}], [\textit{wing}]'', ``\textit{yellow belly and wings}''$\rightarrow$``[\textit{yellow belly}], [\textit{wings}]'', where the adjective word is wrongly categorized. 
To cope with these problems, we leverage two state values $f_1,f_2\in\{0,1\}$ to assist the attribute separation from sentence based on the ``noun-adjective'' rule. 
We have $f_1=1$ for noun word and $f_2=1$ for adjective word. 
If the next word of a noun is ``\textit{has}'' and `` \textit{with}'', then $f_1=0$.
If the word is a conjunction and the previous word and the next word are both adjectives, then $f_2=0$.
If and only if $f_1*f_2=1$, all nouns and adjectives that have been traversed are classified as one attribute.
By this simple rule, we can effectively divide $\mc{S}$ into $M$ attributes, \ie, $\hat{\mc{S}}=\{\mc{A}_{1},\cdots, \mc{A}_{M}\}$.
We compare several sentence parsing methods with the proposed rules with several multi-attribute sentence.
As shown in Tab.~\ref{tab:attr}, the parsing results of our work can effectively extract editable attributes against other related methods.

\begin{table}[t]
	\caption{Comparisons of attribute extraction between the present methods and ours.}	
	\vspace{-10px}
	\begin{center}
		\resizebox{\linewidth}{!}{
	\begin{tabular}{l|p{9.5cm}|l}
			\textbf{Query Sentence}	&a grey bird with webbed feet, a short and blunt orange bill, grey head and wings and has white eyes, a white stripe behind its eyes and white belly and breast & the bird is black with a white belly and an orange bill  \\ 
			\hline\hline
			\textbf{Method} & \textbf{Attribute Extraction}  & \textbf{Attribute Extraction} \\
			\hline\hline
			\textbf{Transformers:~\cite{nikzad2021phraseformer}}	& [`grey bird', `feet', `short', `blunt orange bill', `grey head', `wings', `white eyes', `white stripe', `eyes', `white belly', `breast']& \multirow{2}{*}{[`bird', `black', `white belly', `orange bill']} \\ \hline
			
			\textbf{Topicrank:~\cite{bougouin2013topicrank}}                	& [`grey bird with webbed feet', `blunt orange bill', `grey head', `wings', `white eyes', `white stripe', `eyes', `breast']& \multirow{2}{*}{[`bird', `orange bill']}\\\hline
			
			\textbf{Spacy:~\cite{huang2019tool}}	& [`orange bill', `grey head wings', `white belly', `white eyes', `white stripe'] & \multirow{2}{*}{[`white belly', `orange bill']} \\ \hline
			
			\textbf{Ours}	& [`grey bird', `webbed feet', `short blunt orange bill', `grey head wings', `white eyes', `white stripe eyes', `white belly breast']&  \multirow{2}{*}{[`bird black', `white belly', `orange bill']} \\\hline

		\end{tabular}
		}	
	\end{center}
	\label{tab:attr}
	\vspace{-20px}
\end{table}

In the real world, the attribute distribution are highly imbalanced.
For example, in CUB dataset~\cite{wah2011caltech}, the attribute about ``belly'' appears 34,899 times, but the attribute about ``eyering'' only appears 3,125 times. 
This phenomenon leads to the poor editing on some kinds of attributes with fewer number.
To this end, we propose to combine attributes randomly to augment data.
Specifically, in the training phase, we randomly combine attributes from $\mc{S}$ to build $\hat{\mc{S}_{1}}=\{\mc{A}_{i}\}_{i\in\mc{C}_1}$ and $\hat{\mc{S}_{2}}=\{\mc{A}_{i}\}_{i\in\mc{C}_2}$, where $\mc{C}_1\cup\mc{C}_2=\{1,\cdots,M\}$.
By randomly combining attributes, we obtain more editing alternatives and each attribute will be learned without the limit of distribution imbalance.
To further study the attribute-specific editing, we design a \textit{contrastive training} strategy to make each kind of attribute trained effectively.

\subsection{Contrastive Training using attention}
In a sentence $\mc{S}$, using different attribute combinations to edit image will yield differnt results.
Contrastive training~\cite{he2020momentum} aims to learn a representation to pull ``positive" pairs in certain metric space and push apart the representation between ``negative" pairs. 
That means, based on this observation, we can impose contrastive training on the network to control the editing difference between two attribute combinations.
We implement our design based on the famous AttenGAN~\cite{xu2018attngan}, which calculates the spatial attention of the image w.r.t each word for text-to-image generation.
However, the proposed CA-GAN is quite different from AttenGAN because of the difference on how to construct attention.
In specific, we construct Contrastive Attention for different attribute combinations in the proposed CA-GAN, and with the contrastive training, each attribute editing can be enhanced.

The generator of CA-GAN (See Fig.~\ref{fig2}) has two inputs, image feature $\mb{v}$ by CNN and attribute combination features $\mb{s}_{1}$ and $\mb{s}_{2}$ by RNN. 
We construct cross-modal attention matrix $\mb{C}\in\mathbb{R}^{c \times hw}$ using Cross-Modal Attention Module (CMAM) a as shown in Fig.~\ref{fig2}(e), where $\mb{{C}}_{i,j}$ is calculated as follows:
\begin{equation}
	\mb{C}_{i,j}=\dfrac{\exp(\mb{{s}}^\top_{j}\mb{v}_{i})}{\sum_{k}\exp(\mb{{s}}^\top_{k}\mb{v}_{i})},
	\label{eq:att}
\end{equation}
where $i\in\{1,\cdots,c\}$ is the channel index, $j\in\{1,2\}$ is the attribute combination index.
Then, we can easy to obtain two attention maps w.r.t the attribute combination $\mc{S}_1$ and $\mc{S}_2$ by 
\begin{equation}
	\hat{\mb{C}}_j=\mb{s}^\top\mb{C}, \forall j \in\{1,2\}.
\end{equation}
The Contrastive Attention contains both spatial and channel attention map. 
Because the attention map represents the area of the image to be edited by  attributes, with the attention matrix, we can get the attended feature from different attribute combination attention maps by the means of Hadamard Product.
Thus, given $\mb{s}_{1}$ and $\mb{s}_{2}$, we can easily get six kinds of attention-image pairs using Eq.~\eqref{eq:att} with the original image $\mb{I}$, the edited image $\hat{\mb{I}}_1$ (from combination $\hat{\mc{S}_{1}}$) and $\hat{\mb{I}}_2$ (from combination $\hat{\mc{S}_{2}}$).
This can be seen in Fig.~\ref{fig2}(b).
The six pairs are denoted as 
\begin{enumerate}[(1)]
	\setlength{\itemsep}{0pt}
	\setlength{\parsep}{0pt}
	\setlength{\parskip}{0pt}
	\item $\mb{I}^+_{1}=\hat{\mb{I}}_{1} \times \mb{\hat{C}}_{1}$: \textbf{positive} sample for the \textbf{first} editing attribute combination; 
	\item $\mb{I}^-_{1}=\hat{\mb{I}}_{1} \times \mb{\hat{C}}_{2}$: \textbf{negative} sample for the \textbf{first} editing attribute combination; 
	\item $\mb{I}^+_{2}=\hat{\mb{I}}_{2} \times \mb{\hat{C}}_{2}$: \textbf{positive} sample for the \textbf{second} editing attribute combination;
	\item $\mb{I}^-_{2}=\hat{\mb{I}}_{2} \times \mb{\hat{C}}_{1}$: \textbf{negative} sample for the \textbf{second} editing attribute combination;
	\item $\mb{I}_\text{ori}^1=\mb{I} \times \mb{\hat{C}}_{1}$: editing areas for the \textbf{first} attribute combination on the original image;
	\item $\mb{I}_\text{ori}^2=\mb{I} \times \mb{\hat{C}}_{2}$: editing areas for the \textbf{second} attribute combination on the original image.
\end{enumerate}


The six attended images are fed to a pretrained vgg-16 and get features $\mb{v}^+_{1}$, $\mb{v}^-_{1}$, $\mb{v}^+_{2}$, $\mb{v}^-_{2}$, $\mb{v}^{1}_\text{ori}$ and $\mb{v}^2_\text{ori}$.
Then, we construct the contrastive loss between different features in the image:
\begin{equation}
\mathcal{L}_\text{diff}=-\log \dfrac{\exp(\cos(\mb{v}^-_{1},\mb{v}^{1}_\text{ori}))}{\sum_{p=1}^{N}\exp(\cos(\mb{v}^+_{2},\mb{v}^{1}_\text{ori})}-\log \dfrac{\exp(\cos(\mb{v}^-_{2},\mb{v}^{1}_\text{ori}))}{\sum_{p=1}^{N}\exp(\cos(\mb{v}^+_{1},\mb{v}^{1}_\text{ori})}.
\end{equation}
Through contrastive training, the generator can learn the distribution of each attribute, and establish an accurate association between attribute and image.
In addition, to preserve text-independent background regions, we build the perceptual loss \cite{johnson2016perceptual} to reduce the randomness in the generation process as
\begin{equation}
\underset{}{\mathcal{L}_\text{per}=\dfrac{1}{c \times h \times w}}\lVert \mb{v}^-_{1}-\mb{v}^-_{2} \rVert_{2}^{2}.
\end{equation}

\subsection{Attribute-level Discriminator}
To encourage generators to edit multiple attributes based on sentences, the discriminator should provide attribute-level training feedback to the generator.
Previous work attempted to use sentence-level discriminator~\cite{dong2017semantic} or word-level discriminator \cite{nam2018text, li2020manigan}, but they cannot establish an exact connection between the image area and each attribute.
For instance, in the sentence of ``\textit{the bird has black wings, a black head and a red belly}", when the ``\textit{black}" attribute is passed through the discriminator, sentence-level discriminators do not provide the exact area of the feature in the image, and word-level discriminators localize to both ``\textit{head}" and ``\textit{wing}" regions. 
In order for the discriminator to provide feedback related to each attribute, we propose to develop the attribute-level discriminator.

Our attribute-level discriminator has two inputs, attribute combination features $\mb{s}$ and image feature $\mb{v}$.
We use $\mb{\Delta}_{j}\in\mbb{R}^{c\times 2}$ represents the correlation between the $j^{th}$ ($j\in\{1,2\}$) attribute combination and the whole image
\begin{equation}
	\mb{\Delta}_{j}= \dfrac{\exp((\mb{{s}}^\top_{j}\mb{v}_{i})^\top\mb{v}_{i})}{\sum_{k=1}^{L}\exp((\mb{{s}}^\top_{j}\mb{v}_{i})^\top\mb{v}_{i})}.
\end{equation}
Next, we sum the attribute-weighted image feature $ \mb{\Delta}_{j} $ at the C dimension to get $ \hat{\mb{\Delta}}_{j} $.


Finally, the attribute-level feedback between $\mb{s}$ and $\hat{\mb{\Delta}_{j}}$ is calculated by Binary Cross-Entropy (BCE) loss as
\begin{equation}
\begin{aligned}
\mathcal{L}_\text{attr}= \sum_j\text{BCE}(\mb{s},\mb{\hat{\Delta}_{j}}).
\end{aligned}
\end{equation}
By calculating BCE loss, the discriminator is able to provide attribute-level training feedback to the generator, thus benefiting the alignment between the different attribute features and visual features in the sentence.

\subsection{Objective Function}
The generator and discriminator are trained alternatively by minimizing both the generator loss $\mathcal{L}_\text{G}$ and discriminator loss $\mathcal{L}_\text{D}$. The generator of the whole network contains unconditional adversarial loss and conditional adversarial loss, contrastive loss $\mathcal{L}_\text{diff}$, perceptual loss $\mathcal{L}_\text{per}$ and text-image matching loss $\mathcal{L}_\text{DAMSM}$.
\begin{equation}
\mathcal{L}_\text{G}=-\frac{1}{2} E_{\hat{\mb{I}}\sim P_\text{G}}[\log(\text{D}(\hat{\mb{I}}))]-\frac{1}{2} \mathbb{E}_{\hat{\mb{I}}\sim P_{G}}[\log(\text{D}(\hat{\mb{I}},\mc{S}))]
+\lambda_{1}\mathcal{L}_\text{diff}+\lambda_{2}\mathcal{L}_\text{per}+\lambda_{3}\mathcal{L}_\text{DAMSM},
\end{equation}
where $\mb{I}$ is the real image sampled from the original image distribution, and $\hat{\mb{I}}$ is the generated image sampled from the training model distribution, $\lambda_{1}, \lambda_{2}$ and $\lambda_{3}$ are hyperparameters controlling different losses.
$\mathcal{L}_\text{DAMSM}=\frac{\exp  (\gamma\mb{R}(\mb{s},\mb{v}))}{\sum_{k=1}^{M}\exp  (\gamma\mb{R}(\mb{s},\mb{v}))}$ is used to calculate the matching score between image and text, where $\mb{R}=(c_{i}^{T}e_{i})/(||c_{i}||||e_{i}||)$, $\gamma$ is the smoothing factor, $c$ denotes the picture feature corresponding to the word, and $e$ denotes the feature of the whole sentence.
The complete discriminator objective is defined as:
\begin{equation}
\begin{aligned}
\mathcal{L}_\text{D}=&-\frac{1}{2} \mathbb{E}_{\mb{I}\sim \text{P}_{data}}[\log(\text{D}(\mb{I}))]-\frac{1}{2} \mathbb{E}_{\hat{\mb{I}}\sim P_{G}}[\log(1-\text{D}(\hat{\mb{I}}))] \\
&-\frac{1}{2} \mbb{E}_{\hat{\mb{I}}\sim \text{P}_\text{G}}[\log(1-D(\hat{\mb{I}},\mc{S}))]-\frac{1}{2} \mbb{E}_{\hat{\mb{I}}\sim \text{P}_\text{G}}[\log(\text{D}(\mb{I},\mc{S}))]+\lambda_{4}\mathcal{L}_{attr},
\end{aligned}
\end{equation}
where $\lambda_{4}$ is the hyperparameters controlling $\mathcal{L}_{attr}$.

\section{Experiments}

\subsection{Dataset and implementation detail}

\noindent
\textbf{Dataset:} 
Our model is evaluated on the Caltech-UCSD Birds (CUB) \cite{wah2011caltech} and MS COCO \cite{lin2014microsoft} datasets, where the query sentences of CUB and COCO are the bird description and image captions provided by themself.
CUB contains 200 bird species with 11,788 images where each has 10 sentence deceptions. We pre-encode the sentence by a pretrained text encoder following AttnGAN \cite{xu2018attngan}. The COCO \cite{lin2014microsoft} dataset contains 82,783 training and 40,504 validation images, each  of which  has 5 corresponding text descriptions including word, phrase and sentence. 
Both the datasets have images to be edited by query sentences with multiple attributes.

\begin{figure}[t]
	\centering
	\includegraphics[width=0.9\linewidth]{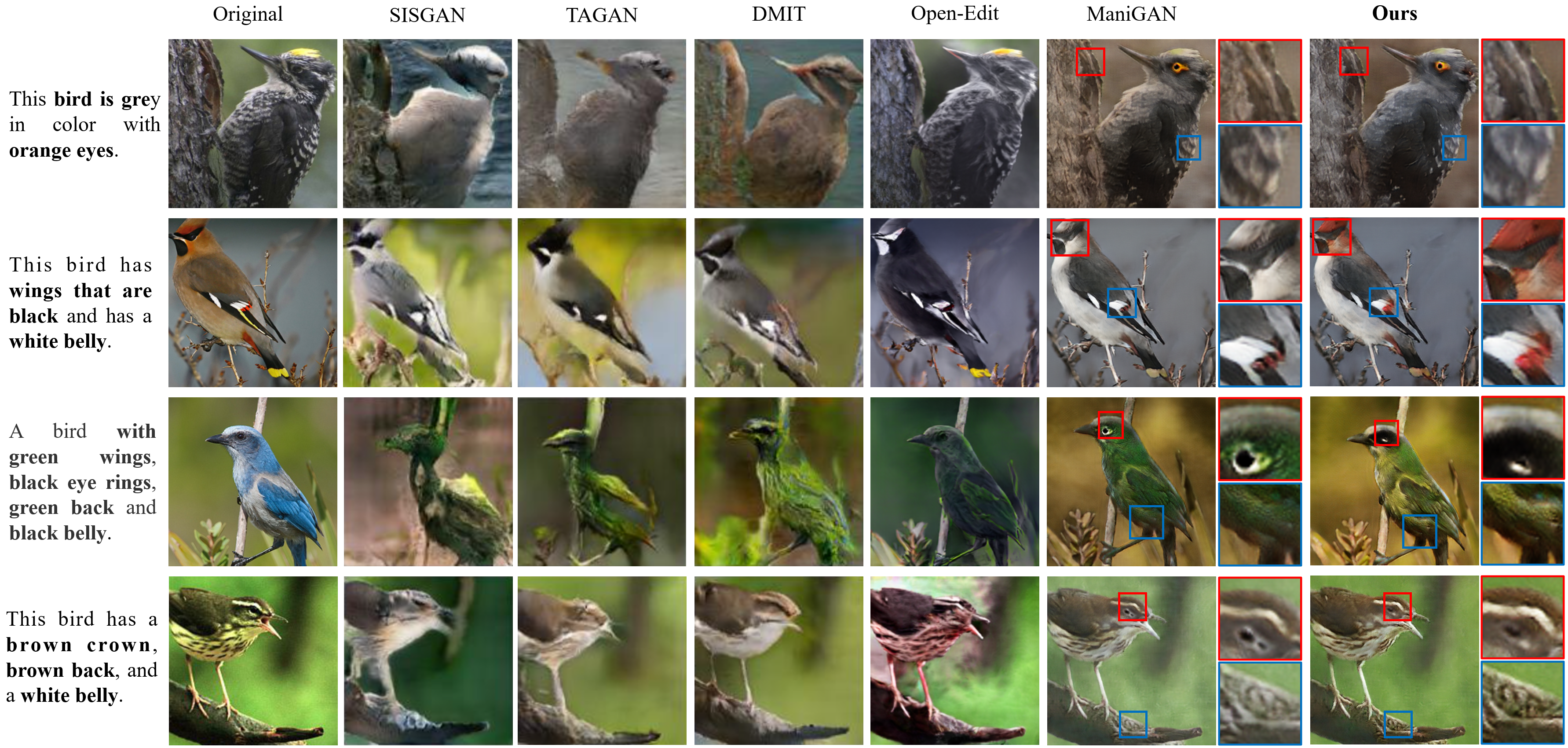}
	\vspace{-10px}
	\caption{
		Editing comparisons on CUB datasets.
	} 
	\label{fig3}
	\vspace{-10px}
\end{figure}

\begin{figure}[t]
	\centering
	\includegraphics[width=0.9\linewidth]{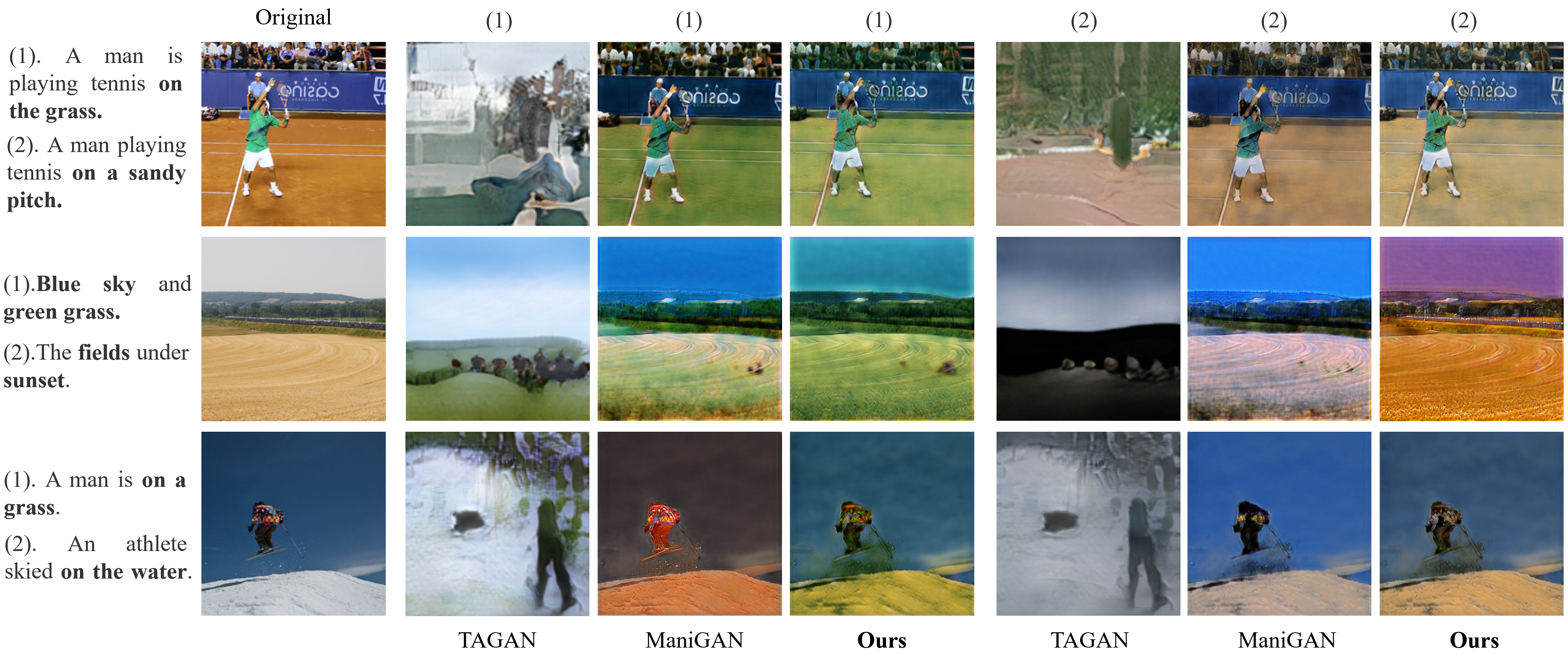}
	\vspace{-10px}
	\caption{
		Editing comparisons on COCO datasets.
	} 
	\label{fig4}
	\vspace{-20px}
\end{figure}

\noindent
\textbf{Implemenation detail.} 
CA-GAN is optimized by Adam~\cite{kingma2014adam} and the learning rate is empirically set to 0.0002. 
The model trains 600 and 120 epochs for CUB and COCO dataset, respectively. 
The hyperparameters $\lambda_{1}$, $\lambda_{2}$, $\lambda_{3}$ and $\lambda_{4}$ are set to 0.7, 0.6, 1 and 0.9 empirically.

\noindent
\textbf{Comparing method.}
The comparing state-of-the-art approaches include SISGAN~\cite{dong2017semantic},   TA-
GAN~\cite{nam2018text}, DMIT~\cite{yu2019multi}, Open-edit~\cite{liu2020open}, and ManiGAN~\cite{li2020manigan}.
Note that these comparing methods never consider the attribute editing but focus on the sentence editing.

\noindent
\textbf{Evaluation Metric.}
We use the Fréchet Inception Distance (FID) \cite{heusel2017gans} and the Learned Perceptual Image Patch Similarity (LPIPS) \cite{zhang2018unreasonable} as the evaluation metrics. 
FID calculates the distance between two multidimensional variable distributions, representing the image generation quality.
LPIPS represents the diversity of the generated images by calculating the L1 distance of the features extracted from AlexNet pre-trained in ImageNet.
We also report the performance by human ranking on each dataset.
We test edited accuracy (Acc.) and realism (Real) by randomly sampling 100 images with the same conditions and collect more than 20 surveys from different population.
Specially, for the COCO dataset, each pair of image and sentence is from the same category. 


\subsection{Comparisons with the state-of-the-arts}

\noindent
\textbf{Qualitative comparison.} Fig.~\ref{fig3} and Fig.~\ref{fig4} show the edit comparisons on the CUB and COCO datasets.
It can be seen that, on  CUB, the three methods, SISGAN, TAGAN, and DMIT, are unable to well generate bird contour information  and  lead to large blurred areas. 
These can be particularly observed in Fig.~\ref{fig4} where the tree trunk color is changed, black eye orbits are lost, and the orange head is modified.
Although ManiGAN shows good image editing performance to some extent, it cannot edit multiple attribute regions better and the background regions change.
We believe that this is due to the fact that ManiGAN is not trained to split and compare attribute features, and thus lacks attribute-level discriminators to offer training feedback related to each feature in the sentence.

\noindent
\textbf{Quantitative comparison.}
As shown in Table~\ref{tab1}, compared with several existing state-of-the-art methods, our method leads to the best FID and LPIPS values on both CUB and COCO, implying that our method is able to achieve high-quality editing images. Moreover, our method also has the highest values for Acc. and Real, indicating that our model generates more favorable images for people.
Our method produces higher quality images.
This means people feel that our editing is better and the images are more realistic.

\begin{table}[t]
	\caption{Quantitative comparison: Fréchet inception distance (FID), Learned Perceptual Image Patch Similarity (LPIPS), Acc. and Real of various methods on CUB and COCO.}	
	\vspace{-10px}
	\begin{center}
	\resizebox{\linewidth}{!}{
	\begin{tabular}{p{30mm}ccccccccc}
		\hline
		& \multicolumn{4}{c}{\textbf{CUB}}                                     & \multicolumn{4}{c}{\textbf{COCO}}          \\ \hline
		\multicolumn{1}{c|}{Method}        & FID $\downarrow$   & LPIPS $\downarrow$ & Acc.(\%) $\uparrow$ & \multicolumn{1}{c|}{Real(\%) $\uparrow$}  & FID $\downarrow$ & LPIPS $\downarrow$ & Acc.(\%) $\uparrow$ & Real(\%) $\uparrow$ \\ \hline
		\multicolumn{1}{c|}{SISGAN~\cite{dong2017semantic}}        & 36.69 & 0.7169   &   5.25      & \multicolumn{1}{c|}{5.3}         & -   &   -   &    -    &   -    \\
		\multicolumn{1}{c|}{TAGAN~\cite{nam2018text}}         & 28.92 & 0.7160   &    9.3     & \multicolumn{1}{c|}{9.7}         &  31.85   &   0.7802    &     8     &      6.2    \\
		\multicolumn{1}{c|}{DMIT~\cite{yu2019multi}}          & 28.01 & 0.7102   &    14.05     & \multicolumn{1}{c|}{16.65}         &   -  &    -   &    -   &   -     \\
		\multicolumn{1}{c|}{Open-edit~\cite{liu2020open}}       & 23.57 & 0.7131   &   16.8      & \multicolumn{1}{c|}{13.85}         &   -  &  -    &    -     &      -   \\
		\multicolumn{1}{c|}{ManiGAN~\cite{li2020manigan}}       & 21.74 & 0.7059   &  24.05       & \multicolumn{1}{c|}{23.5}         &   25.96  &   0.7637    &     42.6     &      44.2   \\
		\multicolumn{1}{c|}{\textbf{Ours}} & \textbf{20.08} & \textbf{0.6893}   &  \textbf{30.4}        & \multicolumn{1}{c|}{\textbf{30.85}}   &   \textbf{24.03}  &  \textbf{0.7438}     &  \textbf{49.4}   &  \textbf{49.6}  \\ \hline
	\end{tabular}
}	
	\end{center}
	\label{tab1}
	\vspace{-10px}
\end{table}

\begin{table}[t]
	\caption{Ablation study quantitative comparison: Fréchet inception distance (FID), Learned Perceptual Image Patch Similarity (LPIPS) on CUB.}
	\vspace{-10px}
	\begin{center}
		\resizebox{.6\linewidth}{!}{
\begin{tabular}{c|c|ccc}
	\hline
	                         &Ours& Ours w/o SPAC & Ours w/o CA & Ours w/o AD \\ \hline
	\multicolumn{1}{c|}{FID} &\textbf{20.08}& 30.17             & 22.36           & 24.32 \\
	\multicolumn{1}{c|}{LPRPS} &\textbf{0.6893}& 0.7153 & 0.7085 & 0.7147 \\ \hline
\end{tabular}
		}	
	\end{center}
	\label{tab2}
	\vspace{-25px}
\end{table}

\begin{figure}[t]
	\centering
	\includegraphics[width=1\linewidth]{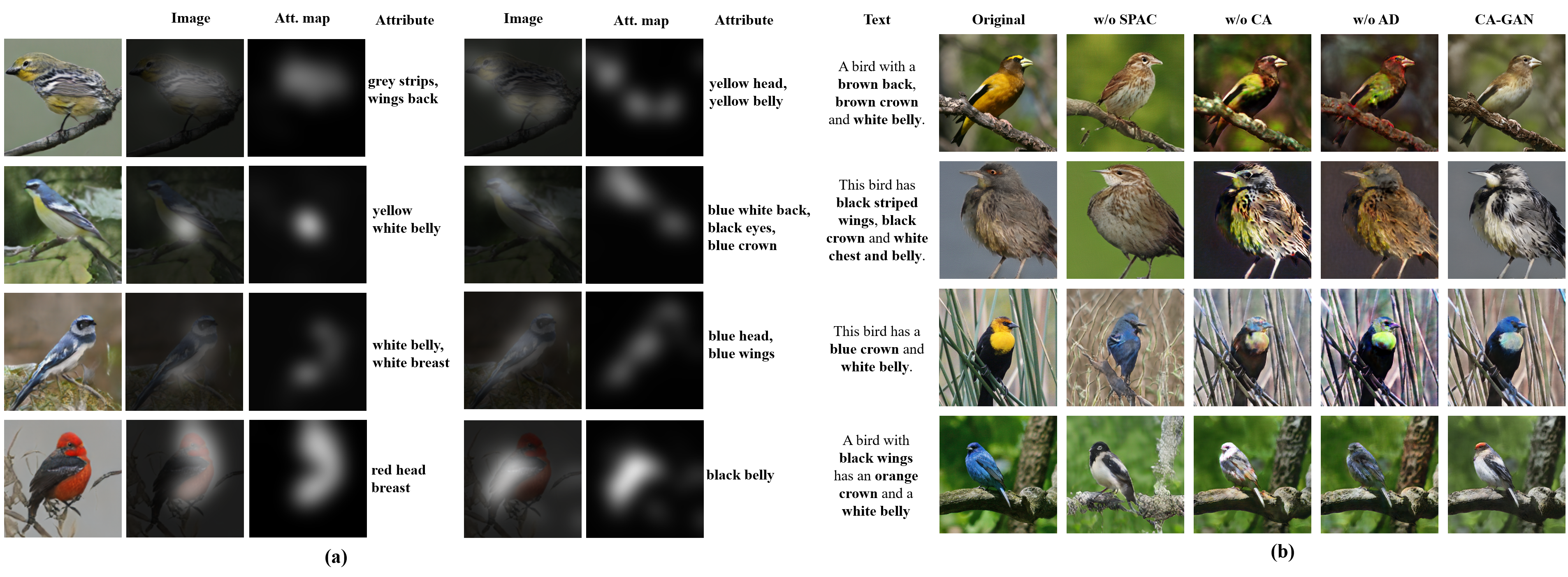}
	\vspace{-20px}
	\caption{
		(a) Visualisation of attention maps. (b) Ablation and comparison studies by removing proposed Sentence Parsing and Attribute Combination (\textbf{w/o SPAC}), removing contrastive attention (\textbf{w/o CA}), removing proposed attribute-level discriminator (\textbf{w/o AD}).
			} 
	\label{fig5}
\end{figure}

\subsection{Ablation Studies}
In this section, we evaluate the main modules in CA-GAN and analyze their impact.
The result can be seen in Fig.~\ref{fig5} (b) and Table~\ref{tab2}.
First, without using the Sentence Parsing and Attribute Combination (w/o SPAC) module, we alternatively combine every two neighboring words in the sentence to construct an editable attribute.
As shown in Fig.~\ref{fig5} (b), in the absence of rules, randomly dividing sentences may perform random editing with no relevant to the query sentence and achieve the worst FID and LPIPS.
Second, we evaluate the effect of Contrastive Attention (w/o CA) module.
We leverage a whole sentences as input, removing the contrastive attention module and contrastive loss.
This confirms our hypothesis that by generating different editable information through sentence parsing and attribute combination, the differences between attributes are amplified by different contrastive attention, which helps the model to focus on the corresponding attributes in a given sentence.
Finally, we evaluate the effectiveness of the proposed Attribute-level Discriminator (w/o AD), and the model cannot effectively operate on the image content based on the sentence information.
For example in Fig.~\ref{fig5} (b), the bird's torso shows blurring and artifacts, and the color of the feathers changes considerably.
This indicates that the generator failed to decompose the different visual attributes due to the lack of attribute-level training feedback, and thus cannot effectively establish attribute-region connections to edit the images.

\subsection{Visualization of contrastive attention and failure cases} 

\begin{figure}[t]
	\centering
	\includegraphics[width=1\linewidth]{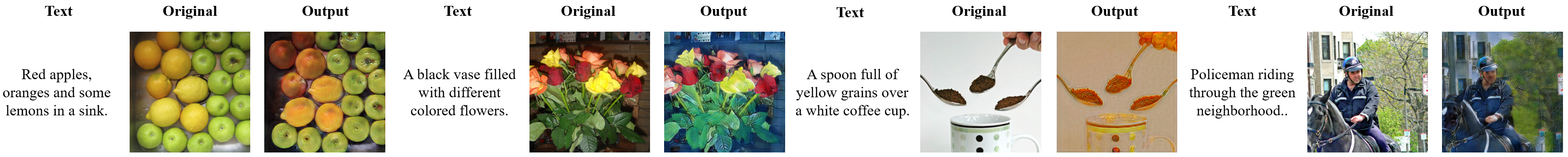}
	\vspace{-20px}
	\caption{
	   Some failure cases on COCO.
	} 
	\label{fig7}
	\vspace{-20px}
\end{figure}

In this section, we visualize the generated results of the attention maps corresponding to the different attributes in the CMAM from the third stage in Fig.~\ref{fig5} (a).
We can observe that the model can better generate the accurate attention maps based on the attribute information after the sentence parsing and attribute combination, and with better accurate position, finer shape, and better semantic consistency between the attributes and the editable content.
We show some failure cases in Fig.~\ref{fig7} on COCO. 
We find that the description semantics maybe fuzzy when there exist multiple categories, and the model may fail to edit the image.

\section{Conclusion}

In this paper, we studied the task of Sentence-based Image Editing.
In contrast to existing methods that cannot produce accurate editing in case of a query sentence with multiple editable attributes, we proposed to enhance the difference between attributes and attained much better performance consequently. Particularly, we developed a novel model called CA-GAN by designing a contrastive attention mechanism  on Generative Adversarial Network. 
We first parsed attributes from the sentence with POS Tagging and generated different attribute combinations.
Then, a contrastive attention module was built to enlarge the editing difference between the combinations.
Last, we constructed an attribute discriminator to ensure the effective editing on each attribute. 
Extensive experiments show that our model can lead to effective editing for sentences with multiple attributes on CUB and COCO datasets. 

\section*{Acknowledgments}
This work was supported by the Natural Science Foundation of China (No. 61876121), Primary Research and Development Plan of Jiangsu Province (No. BE2017663), Natural Science Foundation of Jiangsu Province (19KJB520054), Graduate Research Innovation Program of Jiangsu Province (SJCX20-1119), Scientific Research Project of School of Suzhou Institute of Trade and Commerce (KY-ZRA1805), National Natural Science Foundation of China(No. 61876155), Jiangsu Science and Technology Programme (Natural Science Foundation of Jiangsu Province) (No. BE2020006-4), Key Program Special Fund in XJTLU (No. KSF-T-06).
\bibliography{egbib}

\section*{Appendix}
This document provides supplementary material for the paper ``Each Attribute Matters: Contrastive Attention
for Sentence-based Image Editing'' published on the British Machine Vision Conference (BMVC) 2021.
In this material, we provide the further discussion and illustration of some details, and show more examples of SIE.

\subsection*{A.1~~~~Evaluation Metrics}
In our experiments, we use the \textbf{Fréchet Inception Distance (FID)} \cite{heusel2017gans} and the \textbf{Learned Perceptual Image Patch Similarity (LPIPS)} \cite{zhang2018unreasonable} as the evaluation metrics. 

The FID of an edited image compared to its origin is evaluated by passing it through a pre-trained Inception-v3~\cite{fu2019dual} and computing the distribution difference on the average pooled features.
$\text{FID}$ can be computed by
\begin{equation}
	\text{FID}_{(\mb{I},\hat{\mb{I}})}
	=\left\lVert \mu_{\mb{I}}-\mu_{\hat{\mb{I}}}\right\rVert_{2}^{2}
	+\text{Tr}\left(\Sigma_{\mb{I}}+\Sigma_{\hat{\mb{I}}}-2\left(\Sigma_{\mb{I}}\Sigma_{\hat{\mb{I}}}\right)^{\frac{1}{2}}\right),
\end{equation}
where $\mu_{\mb{I}}$ and $\mu_{\hat{\mb{I}}}$ represents the feature mean of the real image and the generated image.
$\Sigma_{\mb{I}}$ and $\Sigma_{\hat{\mb{I}}}$ represents the covariance matrix of the features of the real image and the generated image.
The smaller the FID value, the closer the distribution between generated image and real image.

We also use LPIPS to calculate the perceptual distance of two images.
Traditionally, Perceptual distance~\cite{zhang2018unreasonable} refers to the visual similarity of two images, the purpose of which is to evaluate the similarity of two images by imitating the human visual senses.
We extract feature from the $l$-th layer and unit-normalize it in the channel dimension, which we designate as $\mb{v}^{l}$ and $\hat{\mb{v}}^{l} \in\mbb{R}^{c\times h\times w}$.
$h_{l}, w_{l}$ is the feature size in different layers.
$\omega_{l}$ is equivalent to computing cosine distance.
$\text{LPIPS}$ can be computed by
\begin{equation}
	\text{LPIPS}_{(\mb{I},\hat{\mb{I}})}=\sum_{l}\frac{1}{h_{l}w_{l}}\sum_{h,w}\lVert \omega_{l}\odot(\mb{v}^{l}-\hat{\mb{v}}^{l})\rVert_{2}^{2}.
\end{equation}

\subsection*{A.2~~~~Sentence Parsing Strategy}
In this paper, we propose a strategy to effectively parse sentence into different attributes, thus to facilitate the subsequent data augmentation and the construction of contrastive learning.

After the POS tagging of sentence $\mc{S}$, we put the corresponding lexical case of each sentence in $\mc{P}$. To classify the different attributes in a sentence, we have the following 5-step strategy.  
1) Screening words for attributes;
2) Determine the adjective attribution of "bird has" case, divide ``bird'' into attributes and set the state of $f_{1}$, $f_{2}$ to 0;
3) If the word is a noun and not bird, the  $f_{1}$ status is set to 1;
4) If the word is an adjective and is not followed by a conjunction, the $f_{2}$ status is set to 1;
5) When $f_{1}$ $\times$ $f_{2}=1$, the attribute is divided.
The detailed algorithm can be seen in Algorithm~\ref{alg1}.
Finally, we get the divided attributes $\hat{\mc{S}}$.


\subsection*{A.3~~~~Discussion of hyperparameters}
The generator and discriminator have trained alternatively by minimizing both the generator loss $\mathcal{L}_\text{G}$ and discriminator loss $\mathcal{L}_\text{D}$. In generator, $\mathcal{L}_\text{diff}$ control different attributes, $\mathcal{L}_\text{per}$ control the invariance of the background, $\mathcal{L}_\text{DAMSM}$ control text-image matching.
In discriminator, $\mathcal{L}_{\text{attr}}$ discriminate the existence of attribute-level information.
\begin{equation}
	\begin{aligned}
		\mathcal{L}_\text{G}=&-\frac{1}{2} E_{\hat{\mb{I}}\sim P_\text{G}}[\log(\text{D}(\hat{\mb{I}}))]-\frac{1}{2} \mathbb{E}_{\hat{\mb{I}}\sim P_{G}}[\log(\text{D}(\hat{\mb{I}},\mc{S}))]\\
		&+\lambda_{1}\mathcal{L}_\text{diff}+\lambda_{2}\mathcal{L}_\text{per}+\lambda_{3}\mathcal{L}_{\text{DAMSM}}
	\end{aligned}
\end{equation}

\begin{equation}
	\begin{aligned}
		\mathcal{L}_\text{D}=&-\frac{1}{2} \mathbb{E}_{\mb{I}\sim \text{P}_{\text{data}}}[\log(\text{D}(\mb{I}))]-\frac{1}{2} \mathbb{E}_{\hat{\mb{I}}\sim P_{G}}[\log(1-\text{D}(\hat{\mb{I}}))] \\
		&-\frac{1}{2} \mbb{E}_{\hat{\mb{I}}\sim \text{P}_\text{G}}[\log(1-D(\hat{\mb{I}},\mc{S}))]-\frac{1}{2} \mbb{E}_{\hat{\mb{I}}\sim \text{P}_\text{G}}[\log(\text{D}(\mb{I},\mc{S}))]\\
		&+\lambda_{4}\mathcal{L}_\text{attr}
	\end{aligned}
\end{equation}

The proposed algorithm is governed by four hyperparameters:
$\lambda_{1}$, $\lambda_{2}$ and  $\lambda_{3}$ are used in the generator to balance the generation of different attributes and to preserve irrelevant backgrounds.
Our model is based on the AttnGAN~\cite{xu2018attngan} model, so for the hyperparameter $\lambda_{3}$, we follow its initial value and do not adjust it.
In the discriminator, $\lambda_{4}$ to control whether each attribute is present in the image or not.
Table~\ref{tb1} shows the sensitivity analysis for hyperparameters using the CUB dataset.
As a rule of thumb, we try from 1 and calculate the FID and LPIPS values for each model.
We found that the models work better when in the range of 0.5 to 1.

\begin{table}[h]
	\centering
	\caption{Hyperparameter analysis.}
	\resizebox*{.4\linewidth}{!}{
		\begin{tabular}{cccccc}
			\hline
			$\lambda_{1}$	&	$\lambda_{2}$	& $\lambda_{3}$ & $\lambda_{4}$  & $\text{FID} \downarrow $  & $\text{LPIPS} \downarrow $\\ \hline
			1	&	1	& 1 & 1 & 23.97 & 0.7085	\\
			0.5	&	0.5	& 1 & 0.5 & 22.75& 0.7073\\
			0.5	&	 0.5	& 1 & 1 & 21.26& 0.7046 \\
			0.5	&	1	& 1 & 1 & 22.09 & 0.7067 	\\
			\textbf{0.7}	&	\textbf{0.6}	& \textbf{1} & \textbf{0.9} & \textbf{20.08}&  \textbf{0.6893} \\
			1.5	&	1.5	& 1 & 1.5 & 24.08 & 0.7091 \\ \hline
	\end{tabular}}
	\label{tb1}
\end{table}

\subsection*{A.4~~~~Additional SIE Examples}
\begin{figure}[h]
	\centering
	\includegraphics[width=0.87\linewidth]{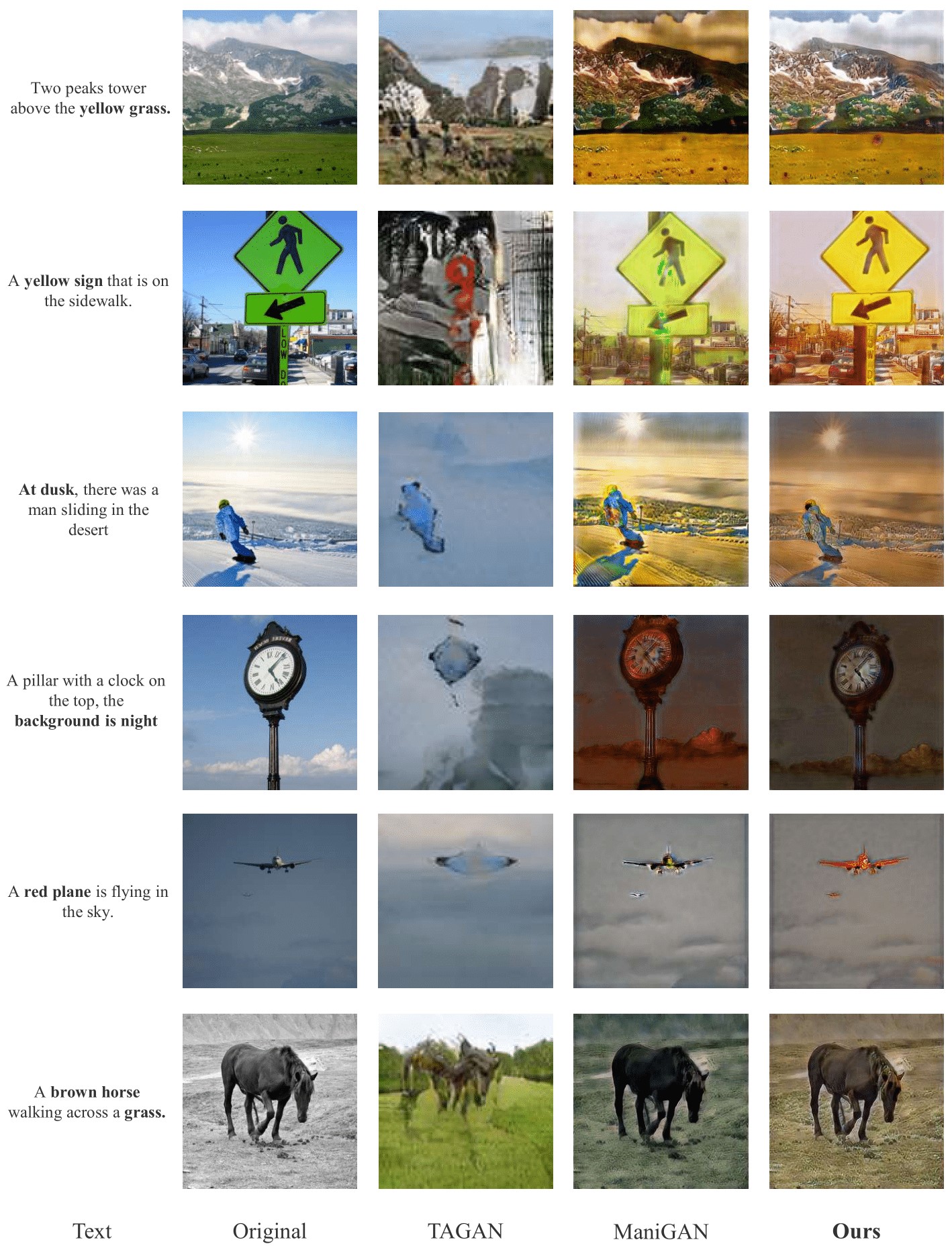}
	\caption{Additional comparison results between TAGAN, ManiGAN, and Ours on the COCO dataset.
	} 
	\label{app_fig1}
	\vspace{-10px}
\end{figure}

\begin{figure}[h]
	\centering
	\includegraphics[width=0.87\linewidth]{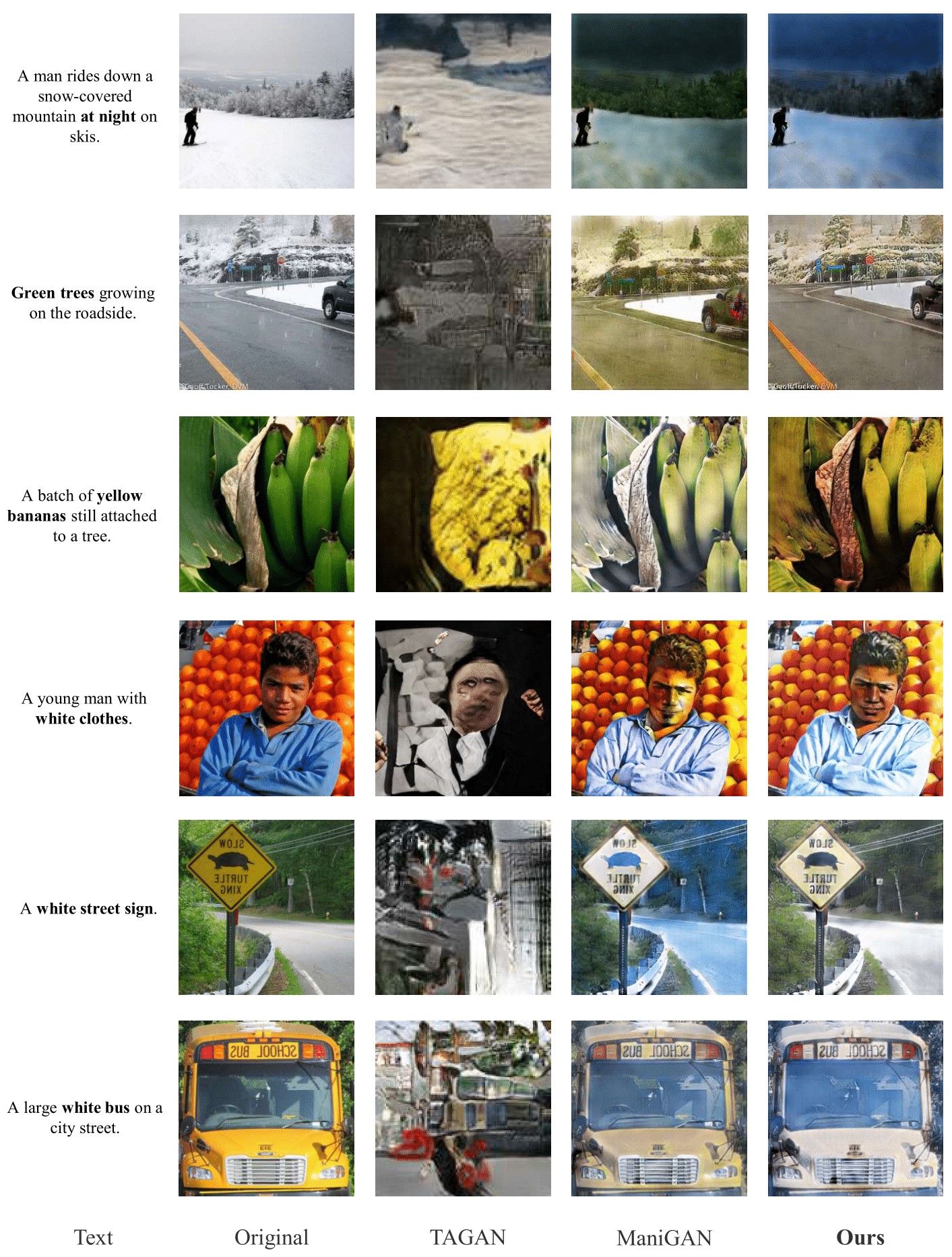}
	\caption{Additional comparison results between TAGAN, ManiGAN, and Ours on the COCO dataset.
	} 
	\label{app_fig2}
	\vspace{-10px}
\end{figure}

\begin{figure}[h]
	\centering
	\includegraphics[width=0.9\linewidth]{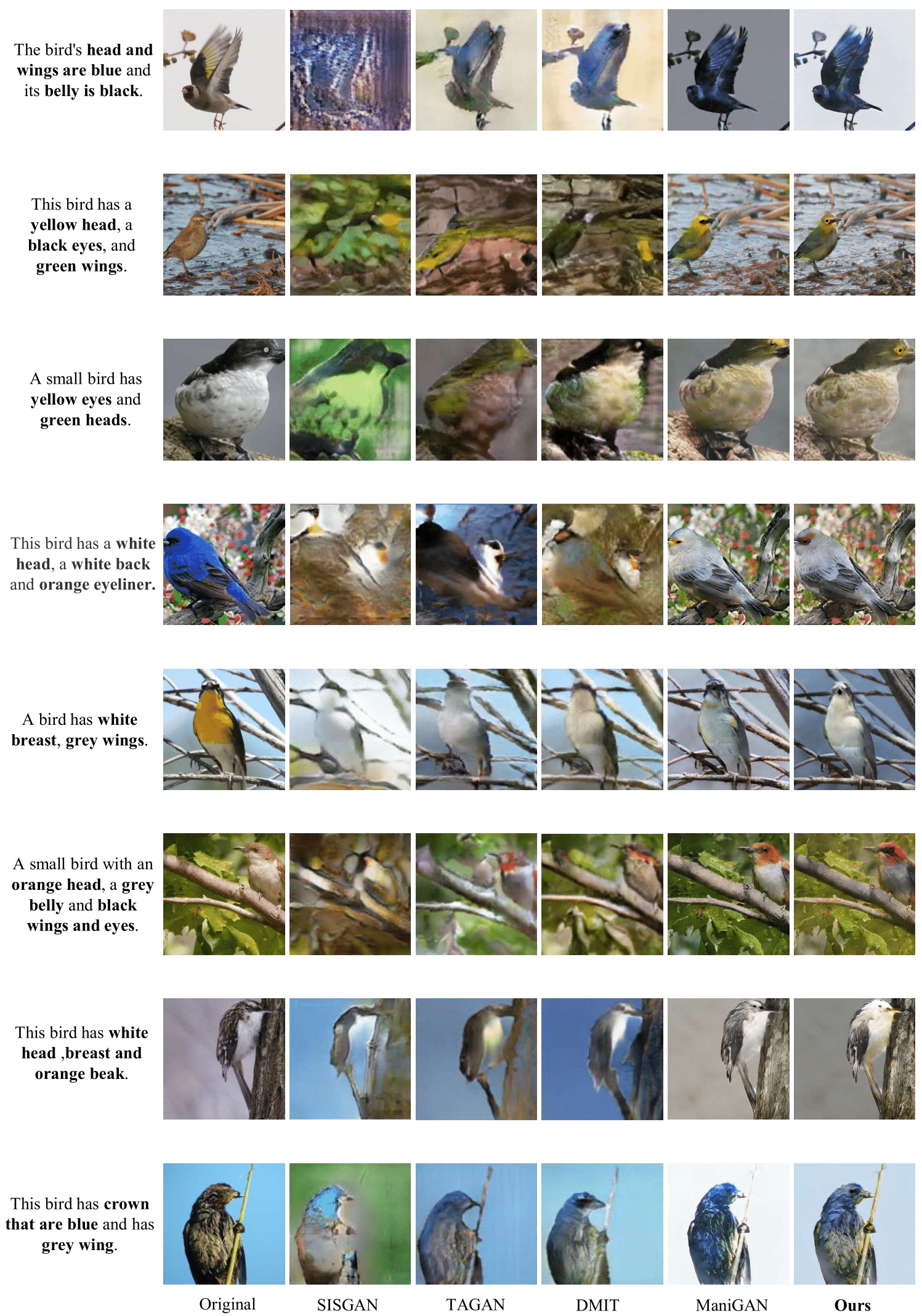}
	\caption{Additional comparison results between SISGAN, TAGAN, DMIT, ManiGAN, and Ours on the CUB dataset.
	} 
	\label{app_fig3}
	\vspace{-10px}
\end{figure}

In Fig.~\ref{app_fig1}~\ref{app_fig2}~\ref{app_fig3}, we show a qualitative comparison of the models on the COCO, CUB dataset.

\begin{algorithm}[t]
	\label{alg1}
	\caption{Sentence Parsing method}
	\KwIn{Sentence $\mc{S}=\{\text{w}_{1},\cdots,\text{w}_{N}\}$, Status control symbols $f_{1}$ =0, $f_{2}$ = 0 , Counters $  n = 0 , m = 0$.}
	\KwOut{Parsed Sentence $\hat{\mc{S}}=\{\mc{A}_{1},\cdots,\mc{A}_{M}\}$, $\mc{A}=\{\text{w}_{i}\}$.}
	$\mc{P}=\{\text{a}_{1},\cdots,\text{a}_{N}\}$ $\leftarrow$ POS tagging ($\mc{S}$);\\
	\For{$ i \leftarrow 0$ \KwTo $len(\mc{S})$}{
		\tcp{\small 1.Screening words for attributes}
		\If{ $\mc{P}_{i}$ $\in$  \rm\{NN, NNS, JJ\} $or$   $\text{w}_{i}$ $\in$ \rm\{``has'', ``with'', ``and''\} }{
			\tcp{\small 2.Judging the adjective attribution in the subject case}
			\If{
				$\rm w_{i}$ = \rm{``bird''} $and$ $\text{w}_{i+1}$ $\in$ \rm\{``has'', ``with''\}}
			{
				$\mc{A}_{n} \leftarrow \{\text{w}_{i},\cdots\}$;\\
				$f_{1}\leftarrow 0,f_{2}\leftarrow $ 0\;
				$n=n+1$;\\
			}
			\tcp{\small 3.Judging the noun case}
			\If{ $\mc{P}_{i}$ = \rm{``NN''} $or$ $\mc{P}_{i}$ $=$  \rm{``NNS''}  $and$ $\text{w}_{i}$ $\neq$  ``bird''}{
				$\mc{A}_{n} \leftarrow \text{w}_{i}$;\\
				$f_{1}\leftarrow$ 1;\\
			}
			\tcp{\small 4.Judging the adjective attribution in conjunctive cases}
			\If{ $\mc{P}_{i}$ $=$  \rm{``JJ''}  and $\text{w}_{i+1}$ $\neq$   \rm{``and''} }{
				$\mc{A}_{n} \leftarrow \text{w}_{i}$;\\
				$f_{2}\leftarrow$ 1;}	
			\tcp{\small 5.Classifying attributes}
			\If{ $f_{1}$ $\times$ $f_{2}=1$}{
				$\mc{A}_{n} \leftarrow \{\text{w}_{i},\cdots\}$;\\
				$f_{1}\leftarrow 0,  f_{2}\leftarrow $ 0\;
				$n=n+1$
			}
		}
		$\hat{\mc{S}}_{m}$ $\leftarrow \mc{A}_{n}$\\
		$m=m+1$	
	}
\end{algorithm}

\end{document}